\documentclass[letterpaper]{article}
\usepackage{aaai16}
\usepackage[utf8]{inputenc}
\usepackage{times}
\usepackage{latexsym,graphicx}
\usepackage{url}
\newcommand{\wc}[1]{}

\newcommand{\ct}[1]{{\it ct: {#1}}}

\newcommand{\ifshort}[1]{}
\newcommand{\iflongelse}[2]{}
\newcommand{\uncite}[1]{}

\newcommand{\edge}[2]{{u\rightarrow{}v}}

\usepackage{balance}
\usepackage{algorithm}
\usepackage{algorithmic}

\newcommand{\bing}[1]{}

\usepackage{multirow}

\title{Distant IE by Bootstrapping Using Lists and Document Structure}

\begin{document}
\author{Lidong Bing$^{\S}$ \ \ Mingyang Ling$^{\S}$ \ \ Richard C. Wang$^{\natural}$ \ \ William W. Cohen$^{\S}$ \\
$^{\S}$Carnegie Mellon University, Pittsburgh, PA 15213\\
$^{\natural}$US Development Center,
Baidu USA, Sunnyvale, CA 94089\\
{ $^{\S}$\{lbing@cs, mingyanl@andrew, wcohen@cs\}.cmu.edu} \\ {$^{\natural}$richardwang@baidu.com}}
\maketitle

\begin{abstract}
Distant labeling for information extraction (IE) suffers from noisy training data.
We describe a way of reducing the noise associated with distant IE by identifying
coupling constraints between potential instance labels.  As one example of coupling,
items in a list are likely to have the same label.
A second example of coupling comes from analysis of document structure: in some corpora,
sections can be identified such that items in the same section are likely to have
the same label.  Such sections do not exist in all corpora, but we show that
augmenting a large corpus with coupling constraints from even a small, well-structured corpus can improve performance substantially, doubling F1 on one task.
\end{abstract}

\section{Introduction}

In distantly-supervised information extraction (IE), a knowledge base
(KB) of relation or concept instances is used to train an IE system.
For instance, a set of facts like
\textit{ad\-verse\-Effect\-Of(melox\-icam, stom\-ach\-Bleed\-ing)},
\textit{int\-eracts\-With(melox\-icam, ibu\-prof\-en)}, might be matched against a corpus, and the
matching sentences then used to generate training data consisting
of labeled entity mentions.  For instance, matching the KB above might
lead to labeling passage \ref{lab:bleeding} from
Table~\ref{tab:meloxicam} as support for the fact
\textit{ad\-verse\-Effect\-Of(melox\-icam, stom\-ach\-Bleed\-ing)}.

A weakness of distant supervision is that it produces noisy training data, when matching errors occur.  E.g., consider using distant learning to classify noun phrases (NPs) into types, like \textit{drug} or \textit{symptom}; matching a polysemous term like \textit{weakness} could lead to incorrectly-labeled mention examples.  Hence distant
supervision is often coupled with learning methods that allow for this
sort of noise by introducing latent variables
for each entity mention (e.g., \cite{hoffmann2011knowledge,riedel2010modeling,Surdeanu:2012:MML:2390948.2391003});
by carefully selecting the entity mentions
from contexts likely to include specific KB facts \cite{wu2010open};
   or by careful filtering of
the KB strings used as seeds \cite{Movshovitz-Attias:2012:BBO:2391123.2391126}.

\iffalse Hoffmann, Raphael, et al. "Knowledge-based weak supervision for information extraction of overlapping relations." Proceedings of the 49th Annual Meeting of the Association for Computational Linguistics: Human Language Technologies-Volume 1. Association for Computational Linguistics, 2011.

Surdeanu, Mihai, et al. "Multi-instance multi-label learning for relation extraction." Proceedings of the 2012 Joint Conference on Empirical Methods in Natural Language Processing and Computational Natural Language Learning. Association for Computational Linguistics, 2012.

bionell Movshovitz-Attias, Dana, and William W. Cohen. "Bootstrapping biomedical ontologies for scientific text using nell." Proceedings of the 2012 Workshop on Biomedical Natural Language Processing. Association for Computational Linguistics, 2012.
\fi

\begin{table}[t]
\hrule
\begin{enumerate}
\item \label{lab:bleeding}  ``Avoid drinking alcohol.  It may increase your risk of stomach bleeding.''
\item \label{lab:adverse} ``Get emergency medical help if you have chest pain, weakness, shortness of breath, slurred speech, or problems with vision or balance.''
\item \label{lab:drug} ``Check the label to see if a medicine contains an NSAID (non-steroidal anti-inflammatory drug) such as aspirin, ibuprofen, ketoprofen, or naproxen.''
\end{enumerate}
\hrule
\caption{Passages from a page describing the drug melox\-icam.} \label{tab:meloxicam}

\end{table}

\begin{figure}[t]
\centerline{\includegraphics[width=0.48\textwidth]{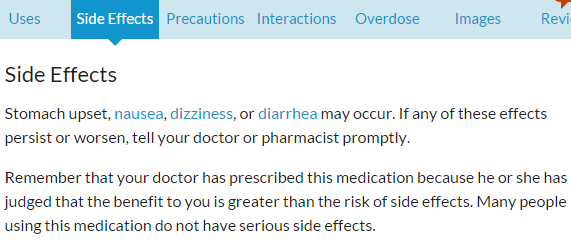}}
%\vspace{-0.1in}
\caption{A structured document in WebMD describing the drug meloxicam.  All documents in this corpora have the same 7 sections.}
\label{fig:side_effect_meloxicam}
\end{figure}

We describe a way of reducing the noise associated with distant IE by identifying coupling
constraints between potential instance labels.  As one example of coupling, NPs in a conjunctive list
are likely to have the same category, a fact used in prior work \cite{bing-EtAl:2015:EMNLP} to propagate NP categories from unambiguous NPs (such as \textit{chest pain} in passage \ref{lab:adverse}) to ambiguous ones (e.g., the mention \textit{weakness} in the same passage). Bing et al. used
propagation methods \cite{ZhuICML2003,DBLP:conf/asunam/LinC10} to
exploit this intuition, by propagating the low-confidence labels associated
with distance supervision matches through an appropriate graph.

\begin{figure*}[t]
\centerline{\includegraphics[width=0.64\textwidth]{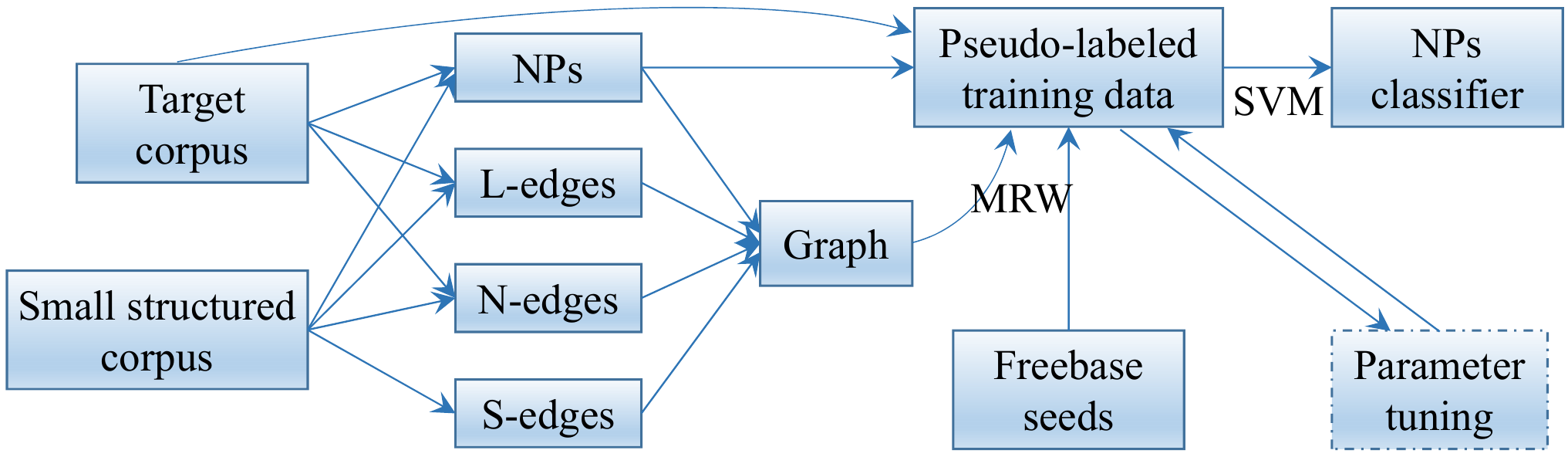}}
%\vspace{-0.1in}
\caption{Architecture of DIEBOLDS}
\label{fig:arch}
\end{figure*}

In this paper we adapt this coupling to extracting relations, rather than NP categories.  We also explore additional types of coupling, derived from analysis of document structure.  In particular, in some corpora, sections can be identified that correspond fairly accurately to relation arguments. For example, Figure~\ref{fig:side_effect_meloxicam} shows part of a small but well-structured corpus (discussed below) which contains sections labeled ``Side Effects".  This document structure cannot be used to directly derive training data (there are many NPs of many types, such as ``doctor" or ``physicial", even in a ``Side Effects" section),
nevertheless, we will show that coupling schemes can be derived and used to improve distantly-supervised IE, even when the test corpus does not contain well-structured sections.

\section{
DIEBOLDS: \textit{\textbf{D}}istant \textit{\textbf{IE}} by \textit{\textbf{BO}}otstrapping using \textit{\textbf{L}}ist and \textit{\textbf{D}}ocument \textit{\textbf{S}}tructure}

Here we describe a pipelined system called DIEBOLDS.  DIEBOLDS parses two corpora: a large \textit{target corpus} and a smaller \textit{structured corpus}, and also perfoms some document analysis on the structured corpus.  It then extracts NP chunks, together with features that describe each NP mention, as well as \textit{coupling information} of various types.  In particular, DIEBOLDS derives edges that define \textit{list coupling}, \textit{section coupling}, and \textit{neighbor coupling}.  DIEBOLDS then creates an appropriate graph, and uses distant supervision, in combination with a
label-propagation method, to find mentions that can be confidently
labeled. From this pseudo-labeled data, it uses ordinary classifier
learners to classify NP mentions by relation types, where the relation indicates the relationship of an NP mention to the entity that is the subject of the document containing the mention.  Extensive experiments are conducted on two corpora, for diseases and drugs, and the results show that this approach significantly improves over a classical distant-supervision approach.  The architecture of the system is shown in Figure~\ref{fig:arch}.

\subsection{Knowledge Base and Corpora}
Distantly-supervised IE is often used to extend an incomplete KB.  Even large curated KBs
are often incomplete: e.g., a recent work showed
that more than 57\% of the nine commonly used attribute/relation values are missing for the
top 100k most frequent PERSON entities in Freebase \cite{west2014knowledge}.
We consider extending the coverage of Freebase in the medical domain, which is currently fairly limited:
e.g., a Freebase snapshot from April 2014 has (after filtering noise with simple rules such as length greater than 60 characters and containing comma) only 4,605 instances in ``Disease or Medical Condition`` type and 4,383 instances in ``Drug'' type, whereas \texttt{dailymed.nlm.nih.gov} contains data on over 74k drugs, and \texttt{malacards.org} lists nearly 10k diseases.

We focus on extracting instances for 8 relations, defined in Freebase, of drugs and diseases.
The targeted drug relations include used\_to\_treat, conditions\_this\_may\_prevent, and side\_effects. The targeted disease relations include treatments, symptoms, risk\_factors, causes, and prevention\_factors.

Our target drug corpus, called DailyMed, is downloaded from
\texttt{dailymed.nlm.nih.gov} which contains 28,590 XML documents,
each of which describes a drug that can be legally prescribed in the United States.
Our target disease corpus, called WikiDisease, is extracted from a Wikipedia dump of May 2015 and it contains 8,596 disease articles.
Large amount of this information in our corpora is in free text. DailyMed includes information about treated diseases, adverse effects, drug ingredients, etc. WikiDisease includes information about causes, treatments, symptoms, etc.

Our corpora are ``entity centric", i.e., each document discusses a single drug or disease. Relation extraction is to predict the type of an entity mention and its relation with the document subject. For instance, the mention \textit{chest pain} is an instance of side\_effects of the drug meloxicam in Table~\ref{tab:meloxicam}.

The structured drug corpus, called WebMD, is collected from \texttt{www.webmd.com}, and each drug page has 7 sections, such as Uses, Side Effects, Precautions, etc. WebMD contains 2,096 pages.
The structured disease corpus, called MayoClinic, is collected from \texttt{www.mayoclinic.org}. The sections of MayoClinic pages include Symptoms, Causes, Risk Factors, Treatments and Drugs, Prevention, etc. MayoClinic contains 1,117 pages.
These sections discuss the important aspects of drugs and diseases, and Freebase has corresponding relations to capture such aspects.

\bing{
\begin{table}[!t]
\center{
\small{
\begin{tabular}{@{}c@{~}|@{~}c@{~}@{~}c@{~}|@{~}c@{~}@{~}c@{}}
  \hline
 & MayoClinic & WikiDisease & WebMD & DailyMed \\
 & (Bootstrap)  & (Target)  & (Bootstrap)  & (Target)  \\
  \hline
  \hline
Documents & 000  & 000&000& 000\\
Unique sentences & 000  & 000&000& 000\\
NPs & 000  & 000&000& 000\\
Lists & 000  & 000&000& 000\\
  \hline
\end{tabular}
%\end{small}
}
}
\caption{Target corpora and bootstrap corpora.\label{t:data}}
\end{table}
}

\subsection{Propagation Graph}

\subsubsection{List Extraction and Graph with List Edges}
We use the GDep parser~\cite{sagae:2007b}, a dependency parser trained on the GENIA Treebank, to parse the corpora. We use a simple POS-tag based noun-phrase (NP) chunker, and extract a list for each coordinating conjunction that modifies a nominal. For each NP we extract features (described below); and for each identified coordinate-term list, we extract its items.

The extracted lists and their items, as well as entity mentions and their corresponding NPs, are used to create bipartite graph. One set of vertices correspond to entity mentions, where each mention is encoded as a pair, consisting of the subject entity for the document, paired with the string corresponding to the NP itself.
The other set of vertex identifiers are for the lists.
A mention not inside a list is regarded as a singleton list that contains only one item.
If an NP is contained by a list, an edge between the NP vertex and the list vertex is included in the graph.
We refer to such edges as list edges (L-edges for short).
An example bipartite graph is given in Figure \ref{fig:graph_section_link} (ignore for now the dashed and dotted links). There are 9 instances of side\_effects relations from three lists and four mentions extracted from two drugs.

\subsubsection{Section Edges}

For each subject entity, there are only a few pages (typically one) that discuss that entity.  Hence a graph containing only list edges is not well-connected: generally the edges only link vertices from a single document. To improve the connectivity, we augment the graph of a target corpus with a graph derived from the  structured corpus, thus building an augmented graph.

\begin{figure}
\centerline{\includegraphics[width=0.42\textwidth]{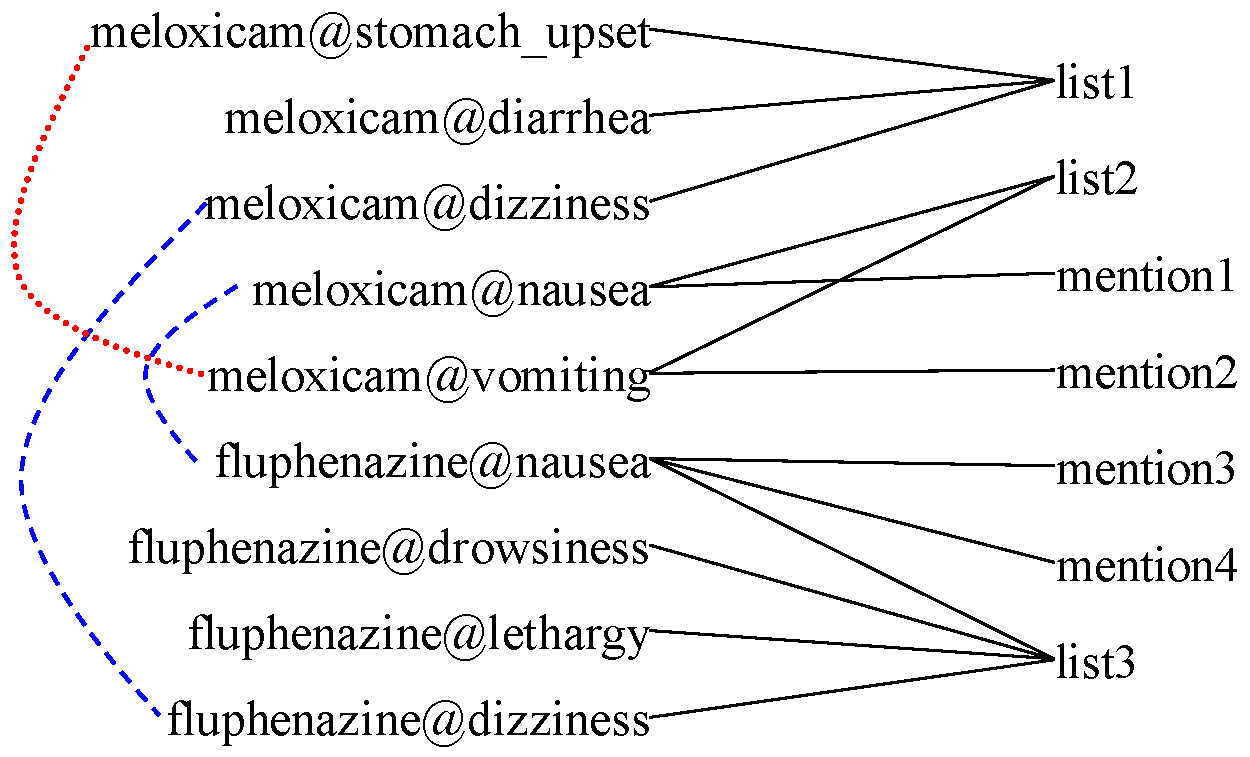}}
%\vspace{-0.1in}
\caption{An example of label propagation graph. }
\label{fig:graph_section_link}
\end{figure}

Firstly, a bipartite graph for a structured corpus is constructed. In addition to lists, we employ the section information of the structured documents: specifically, edges are added between the drug-NP pairs (or disease-NP pairs) as exemplified by the dotted and dashed links in Figure \ref{fig:graph_section_link}.
We add a dashed blue edge for two drug-NP pairs if their NP strings match and the two NPs come from the same section of two documents. In Figure \ref{fig:graph_section_link}, two such edges are added because of the same section ``Side Effects'' in two drug documents, i.e., meloxicam and fluphenazine.
The intuition is that if an NP appears in the same section of different documents, the occurrences are very likely to have the same relation with the corresponding document subjects. Also, if two NPs appear in the same section of a document, they might have the same relation
with the document subject. For instance, both ``vomiting'' and ``stomach\_upset'' appear in ``Side Effects'' section of meloxicam, it is reasonable to infer they might have the same relation label.
We refer to those edges as section edges (S-edges).

Our target corpora have thousands of different section titles, many of which are not related in any way to the relations being extracted, so we do not add S-edges for their sections.
Now we augment the bipartite graph of a target corpus with the graph for a corresponding structured corpus.

\subsubsection{Neighbor Edges}

We might want to further link the drug-NP (or disease-NP) pairs of a target corpus
and a structured corpus with similarity-based edges. We add a weighted edge for
two drug-NP pairs if their NP mentions are similar, where the weight
is in $(0, 1]$ and calculated with TFIDF-weighted BOW of contexts for the NPs. The context contains all
words, excluding NP itself, from sentences containing the NP. We weight these edges with the cosine similarity of the two BOW objects, after TFIDF weighting. (Note that
that the weight of other edges is 1.)
Such weighted edges capture the intuition that if two NPs have similar
contexts, they are very likely to have the same relation label.
We refer to such edge as near-neighbor edges (N-edges).

Obviously, if both drug names and NP strings in two drug-NP pairs
match, they are merged as the same node in the augmented graph.
For NP string and drug/disease name matching, we employ SecondString with
SoftTFIDF as distance metric \cite{CohenIIWeb2003} and the match threshold is 0.8.
It is also used for all baselines compared in the Experiments section.

\subsection{Label Propagation}
Considering the nature of those added edges, it seems plausible to use label propagation on the above graph to propagate relation types from seed drug-NP (disease-NP) pairs with known relation types, e.g., those matching the triples in KB, to more pairs and lists across the graph.

This can be viewed as semi-supervised learning (SSL) of the pairs that may denote a relation (e.g., ``used\_to\_treat'' or ``side\_effects'').
We adopt an existing multi-class label propagation method,
namely, MultiRankWalk (MRW) \cite{DBLP:conf/asunam/LinC10},
to handle our task, which is a graph-based SSL related to personalized
PageRank (PPR) \cite{Haveliwala03ananalytical} (aka random walk with restart  \cite{DBLP:conf/icdm/TongFP06}).
Given a graph represented by matrix $S$,
a vector probability
distribution over the nodes $\mathbf{v}$ is found that satisfies the equation
$$\mathbf{v}=\alpha\mathbf{r}+(1-\alpha)SD^{-1}\mathbf{v}$$
where $D=\Sigma_i{S_i}$, $SD^{-1}$ is the column-stochastic transition
matrix of the graph; and $\mathbf{r}$, the \emph{seed vector}, is a uniform distribution over the
labeled training instances of each class (here the facts from KB); and
$\alpha$ is the restart probability.  (In the experiments we use $\alpha=0.1$.)
The vector $\mathbf{v}$ can be interpreted as the probability
distribution of a random walk on the graph, where at each step there is
a probability $\alpha$ to ``teleport'' to a random
node with distribution $\mathbf{r}$. MRW performs one such computation of a vector
$\mathbf{v}_c$ for each class $c$,
then assigns each instance $i$ to the class $c$ with highest score, i.e.
it predicts for $i$ the label $c=\textrm{argmax}_c \mathbf{v}_c(i)$.

MRW can be viewed as simply computing one personalized PageRank vector for each class, where each vector is computed using a personalization vector that is uniform over the seeds, and finally assigning to each node the class associated with its highest-scoring vector.  MRW's final scores depend on centrality of nodes, as well as proximity to the seeds, and in this respect MRW  differs from other label propagation methods (e.g., \cite{ZhuICML2003}): in particular, \emph{it will not assign identical scores to all seed examples}.  Hence MRW will weight up seeds that are well-connected to other seeds, and weight down seeds that are in``outlying" sections of the graph.
The MRW implementation we use is based on ProPPR \cite{wang2013programming}.

\subsection{Classification}

One could imagine using the output of MRW to extend a KB directly. However, the process described above cannot be used conveniently to label new documents as they appear.  Since this is also frequently a goal, we use the MRW output to train a classifier, which can be then used to classify the entity mentions (singleton lists) and coordinate lists in any new document, as well as those not reached ones in the above graph.

We use the same feature generator for both mentions and lists.  Shallow features include: tokens in the NPs, and character prefixes/suffixes of these tokens; tokens from the sentence containing the NP; and tokens and bigrams from a window around the NPs.  From the dependency parsing, we also find the verb which is the closest ancestor of the head of the NP, all modifiers of this verb, and the path to this verb.  For a list, the dependency features are computed relative to the head of the list.

We used an SVM classifier \cite{CC01a} and discard singleton features, and also the most frequent 5\% of all features (as a stop-wording variant).
%In practical application, we will inevitably encounter a lot of ``other'' lists, i.e.,
%not belonging to the interested instance types in this task.
%However, we do not have other examples for training such a classifier.
We train a binary classifier on the top N lists (including mentions and coordinate lists) of each relation, as scored by MRW. A linear kernel and defaults for all other parameters are used.
If a new list or mention is not classified as positive by all binary classifiers, it is predicted as ``other''.

\subsection{Parameter Tuning}

\wc{not quite clear how the freebase holdouts are used here....}
Two important parameters in DIEBOLDS are the seed number for label propagation
with MRW and the top N number for generating training examples of SVM.
Here we describe our method for tuning them.
The evaluation data is generated with a validating set of facts.
Specifically, these facts are used as seeds for MRW and the
top 200 lists (singleton and coordinate lists)
of each relation, as scored by MRW, are collected. We regard these lists as pseudo-labeled
examples to test the performance of trained classifiers in DIEBOLDS.
Their feature vectors are generated in the same way as above.
We refer to total available seeds for DIEBOLDS as development set,
no overlapping with the validating set here.

\begin{figure}
\centerline{\includegraphics[width=0.44\textwidth]{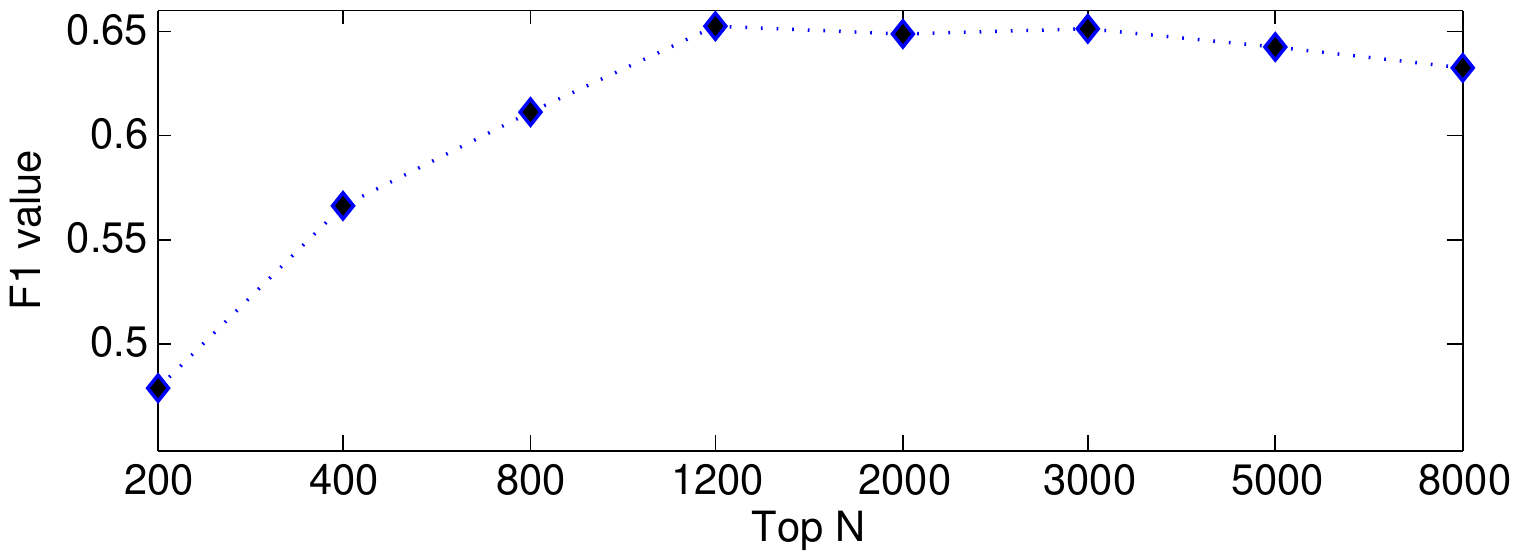}}
%\vspace{-0.1in}
\caption{Effect of top N number, for generating training examples of SVM, on classification performance. }
\label{fig:top_seed100}
\end{figure}

The effect of top N number when using 100\% of development seeds is given in Figure \ref{fig:top_seed100}.
As we expected, too few examples or too many examples are not effective for training
an accurate classifier. The reason is that, if the examples are too few, they
are not adequate to train a classifier with good generalization capability.
One the other hand, if N is too large, the quality of the involved examples
cannot be guaranteed, which also degrades the accuracy of the trained classifier.
A good aspect can be observed from Figure \ref{fig:top_seed100} is that the classifier's
performance is quite stable in a large N range, from 1,200 to 5,000. It indicates that
DIEBOLDS is quite robust and its classification performance is not very sensitive to this parameter.

\begin{figure}
\centerline{\includegraphics[width=0.44\textwidth]{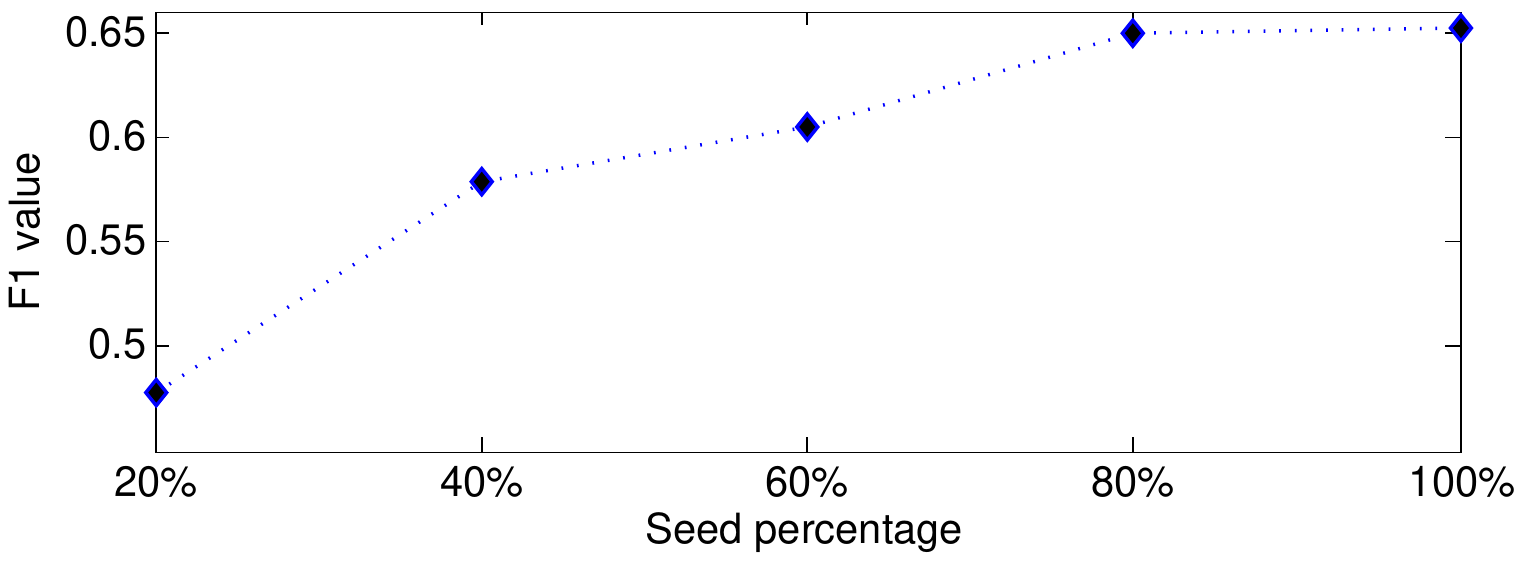}}
%\vspace{-0.1in}
\caption{Effect of the seed number, for label propagation with MRW, on classification performance. }
\label{fig:seed_top1200}
\end{figure}

We also try different ratios of the development set as seeds
of label propagation in DIEBOLDS.
The results are given in Figure \ref{fig:seed_top1200}.
When the seed number is mall, say 20\%, the trained classifier is
not effective. As the seed number increasing, F1 value gets improved.
The values of 80\% and 100\% are quite similar, and it shows a certain number of seeds will almost
achieve the best result, and the marginal improvement with more seeds
is limited. It is because the label propagation with MRW helps collect
sufficiently good training examples with less number of seeds, (i.e. 80\%).

\bing{F1 values in Figures \ref{fig:seed_top1200} and Figure \ref{fig:top_seed100} are much higher
than those in Table \ref{t:prf}, because the testing examples here are from the top
ranked ones from MRW and they are easier to predict. This way of tuning parameters
might not be the optimal solution, nevertheless it is a workable way because to label
enough pages for tuning is a labor-intensive work.}

\bing{
In our implementation, the used classification model is support vector machine (SVM) \ct{xxx}.
Before training the SVM model, the features are first filtered with simple rules. If a feature only appear
once among the training examples, it is filtered. After that, we remove the
top frequent features by a certain percentage for different features.
Specifically, for list-BOW, prefix/suffix, sentence-BOW, list-context, and
dependency features are 0\%, 5\%, 10\%, 10\%, and 10\%.
We do not remove the frequent list-BOW features, it is because the most frequent ones
are usually the instances of some type, such as ``vomiting'' and ``nausea''.
A larger fraction from the top of sentence-BOW, list-context, and
dependency features are removed, because these features are very noisy, containing
stopwords, measurement units, etc.
LibSVM  with a linear kernel is employed for training, and the other parameters are defaulted. }

\section{Experiments}

\subsection{Evaluation Datasets \footnote{We released some data at
http://www.wcohen.com.}}
For the first dataset, we manually labeled 10 pages from WikiDisease corpus and
10 pages from DailyMed corpus. The annotated text fragments
are those NPs that are object values of those 8 relations, with the
drug or disease described by the corresponding document as the relation
subject. In total, we collected 436 triple facts for disease domain
and 320 triples facts for drug domain.
A pipeline's task is to extract the objects of the
relations in a given document.

For the second dataset, we employ questions in the training dataset of BioASQ3-Task
B \footnote{http://participants-area.bioasq.org/general\_information/Task3b/}
to examine the ability of DIEBOLDS output on answering those questions.
This dataset contains four types of questions: yes/no questions, factoid questions,
list questions, and summary questions \cite{bioasq}. We focus on
factoid and list questions because these questions require a particular entity
name (e.g., of a disease, drug, or gene), or a list of them as an answer.
We only keep the questions that are related to the relations in this paper, and finally we get 58 questions, including 37 factoid questions and 21 list questions.
Each natural language question is translated to a structured database query,
which can be evaluated on any KB. For instance, the answer of query(treatsDisease, daonil, Y)
is expected to be a disease name, i.e. ``diabetes mellitus''.

\subsection{Baselines}
The first two baselines are distant-supervision-based.
A DS baseline attempts to classify
each NP in its input corpus into one of the interested
relation types or ``other'' with the training seeds as distance
supervision. Each sentence in the corpus is processed with the
same preprocessing pipeline to detect NPs. Then,
these NPs are labeled with the training seeds.
The features are defined and extracted in the
same way as we did for DIEBOLDS, and binary
classifiers are trained with the same method.
The first DS baseline only generates labeled
examples from the target corpus, and it is named \textbf{DS1}.
While the second DS baseline uses both target corpus and structured
corpus, and it is named \textbf{DS2}. The third baseline applies list structure
into DS1, and it first preforms label propagation with MRW on the
bipartite graph of target corpus.
Then binary classifiers are trained with the top N lists
scored by MRW in the same way. This baseline is
named \textbf{DS+L}.

\subsection{Variants of DIEBOLDS}
We also investigate different variants of DIEBOLDS. The first variant removes S-edges
and N-edges from the graph of DIEBOLDS when propagating labels,
and it is named DIEBOLDS-SN. By removing S-edges and N-edgess respectively,
we have two more variants, named DIEBOLDS-S and DIEBOLDS-N.
We use the same way to tune the parameters for these variants and also the baseline DS+L.
Specifically, all baselines and variants employ 100\% of the training seeds, and
top N values are determined as the ones achieved the best performance on the tuning examples.

%\textbf{List coupling baselines}.
%These baselines examine the results of only using list structure to coupling source. The first list coupling baseline only conducts label propagation on the graph built with the target corpus, and it is called LC1. The second one conducts
%label propagation on the graph built from the merged data of target corpus and structured corpus, and it is called LC2.

%\textbf{Soft edge coupling pipelines}.
%One might want to like the drug-NP (or disease-NP) pairs of two corpora with similarity-based edges. In fact, we also thought this might be a good strategy to further enhance the pipeline. Therefore, we also show two pipelines with such coupling.
%We add a weighted edge for two drug-NP pairs if their NP strings match and the weight is in $(0, 1]$ and calculated with TFIDF-weighted BOW contexts of NPs in their documents. Such context is built with words, excluding NP itself, from sentences containing. The weight of other edges is 1 in all graphs. By adding such weighted edges to the graph of LC2 and the graph of DIEBOLDS (having list couping and section coupling), we get two soft edge coupling pipelines, named SC1 and SC2, respectively.

\subsection{Experimental Settings}

\wc{is this the total number of seeds or total number of matches?}

We extracted triples of these 8 target relations from Freebase. Specifically,
if the subject of a triple matches with a drug or disease name
in our target corpora and its object value also appear in that document,
it is extracted as a seed.
For disease domain, we get 1,524, 1,976, 593, 674, and 99
triples for treatments, symptoms, risk\_factors, causes, and
prevention\_factors, respectively. For drug domain, we get 2,973, 229, and 243
triples for used\_to\_treat, conditions\_his\_may\_prevent, and
side\_effects, respectively. These triples are split into development
set and validating set in the ratio of 9:1. The development set is used as seed of MRW,
and the validating set is used to validate different parameters.

Note that we did not use the seeds that only match with the structured corpora, because we aim
at examining the effect of using structured corpora on the extraction of target corpus
and excluding such seeds will avoid the bias because of more seeds.
We report the average performance of 3 runs, and each
run has its own randomly generated development set and validating set
 to avoid the bias of seed sampling,

%\subsection{Results of Recovering KB}

\subsection{Results on Labeled Pages}
DS+L and DIEBOLDS variants can classify both NPs and coordinate lists.
After that, lists are broken into items, i.e. NPs, for evaluation.
We evaluate the performance of different pipelines
from IR perspective, with a subject (i.e., document name)
and a relation together as a query, and extracted NPs as retrieval results.
Thus, we have 50 and 30 queries for disease domain and drug domain, respectively.
The predicted probability by the binary classifiers serves as the ranking score inside each query.

\begin{table}[!t]
\center{
\begin{tabular}{@{}c@{~}|@{~}c@{~}@{~}c@{~}@{~}c@{~}|@{~}c@{~}@{~}c@{~}@{~}c@{}}
  \hline
 &  & Disease &  &  & Drug & \\
\cline{2-7}
& P &    R   & F1 & P & R & F1 \\
\hline
  \hline
DS1 & 0.117 &  0.350 &  0.175 & 0.020 & 0.268 & 0.037 \\
DS2 & 0.115 &  0.361 &  0.174 & 0.018 & 0.254 & 0.034 \\
DS+L & 0.122 &  0.380 &  0.184 & 0.031 & 0.432 & 0.057 \\
Freebase & 0.202 &  0.037 &  0.062 & 0.318 & 0.022 & 0.041 \\
\hline
DIEBOLDS-SN & 0.128 &  0.374 &  0.191 & 0.045 & 0.451 & 0.082 \\
DIEBOLDS-S & 0.136 &  \textbf{0.382} &  0.198 & 0.048 & \textbf{0.480} & 0.088 \\
DIEBOLDS-N & 0.131 &  0.372 &  0.194 & 0.047 & 0.419 & 0.085 \\
\hline
DIEBOLDS & \textbf{0.143} &  0.372 &  \textbf{0.209} & \textbf{0.050} & 0.435 & \textbf{0.090} \\
\hline
\end{tabular}
%\end{small}
}
\vspace{-0.2cm}
\caption{Comparison between baselines and DIEBOLDS on extraction results of the labeled pages.\label{t:prf}}
\vspace{-0.4cm}
\end{table}

The results evaluated by precision, recall and F1 measure are given in Table \ref{t:prf}.
DIEBOLDS and its variants outperform the baselines in all metrics.
DIEBOLDS is the most effective pipeline in both domains, and its
improvement over DS1, pure distant supervision, on F1 is about 20\%
in disease domain, and more than 100\% in drug domain.

The precision of DIEBOLDS is consistently better than its variants. Presumably, this is true because
as more linking information is added into the graph, the top-scored
lists or NPs in MRW are becoming less noisy.
DIEBOLDS-S achieves the best recall values in both domains. By removing the S-edges
from DIEBOLDS, N-edges become more important in the graph
and MRW walks to more diverse NPs and lists. Thus, the trained classifier
has better generalization capability and achieves better recall values.
On the other hand, its precision is affected.
DIEBOLDS-SN outperforms DS+L under most metrics in both domains.
Both of them explore list information in label propagation, but the
difference is that DIEBOLDS-SN employs a merged graph of target corpus and
structured corpus. Thus, the lists from structured corpus enhances the transduction capability
of the graph.

The performance order of different pipelines is very stable. The more
information of list and document structure is used, the better the performance is.
It shows that these types of information are all value-added and
combining them is a workable way to get better results. Without using the structured corpus, DS+L still achieves encouraging improvements over DS1
list is a useful resource by itself.
 One interesting
observation is that although DS2 also uses the distantly labeled
examples in the structured corpus, its performance is similar or even worse
than DS1. It shows that simply adding some examples from another corpus is not an effective approach to upgrade the performance.
The results in drug domain are much lower than those of disease domain.
The main reason is that documents of DailyMed are usually quite long, and
too much description of different aspects of a drug overwhelms the
targeted facts.

We also employ Freebase as a comparison system, and use its facts as the system output.
It is not unexpected to observe very low recall values in both domains, since the coverage
of Freebase on specific domains such as biomedical domain is limited.
Specifically, Freebase only contains 16 disease triples and 7 drug triples of those annotated ones.
However, the precision values are also not quite high, especially for disease domain.
The reason is two-fold. First, our labeled pages do not contain all facts of those relations,
but Freebase does contain some facts of those missing ones from the labeled pages.
The second reason is that Freebase is not noise-free, and some facts in it are actually wrong.

\begin{figure}
\centerline{\includegraphics[width=0.45\textwidth]{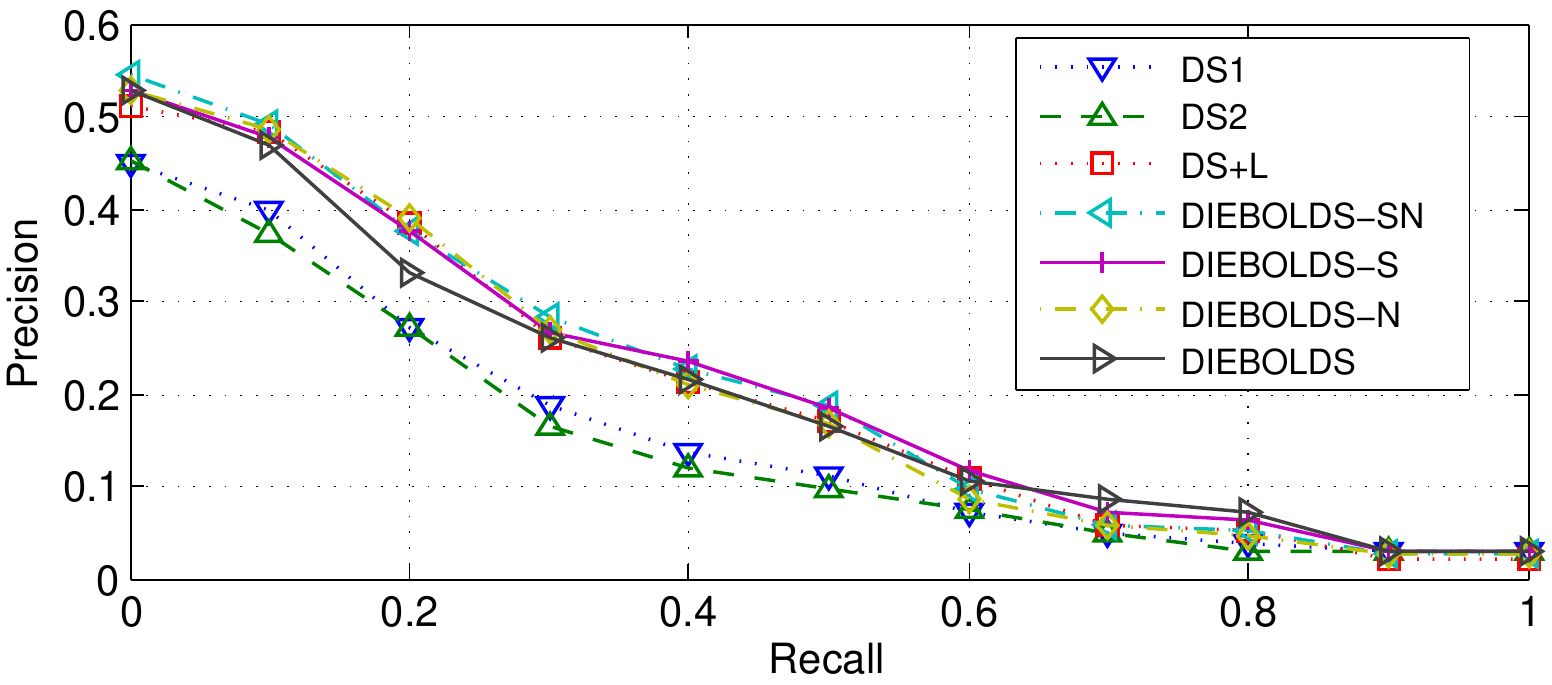}}
%\vspace{-0.1in}
\caption{Precision-recall curve of disease domain. }
\label{fig:pr-curve-disease}
\end{figure}

\begin{figure}
\centerline{\includegraphics[width=0.45\textwidth]{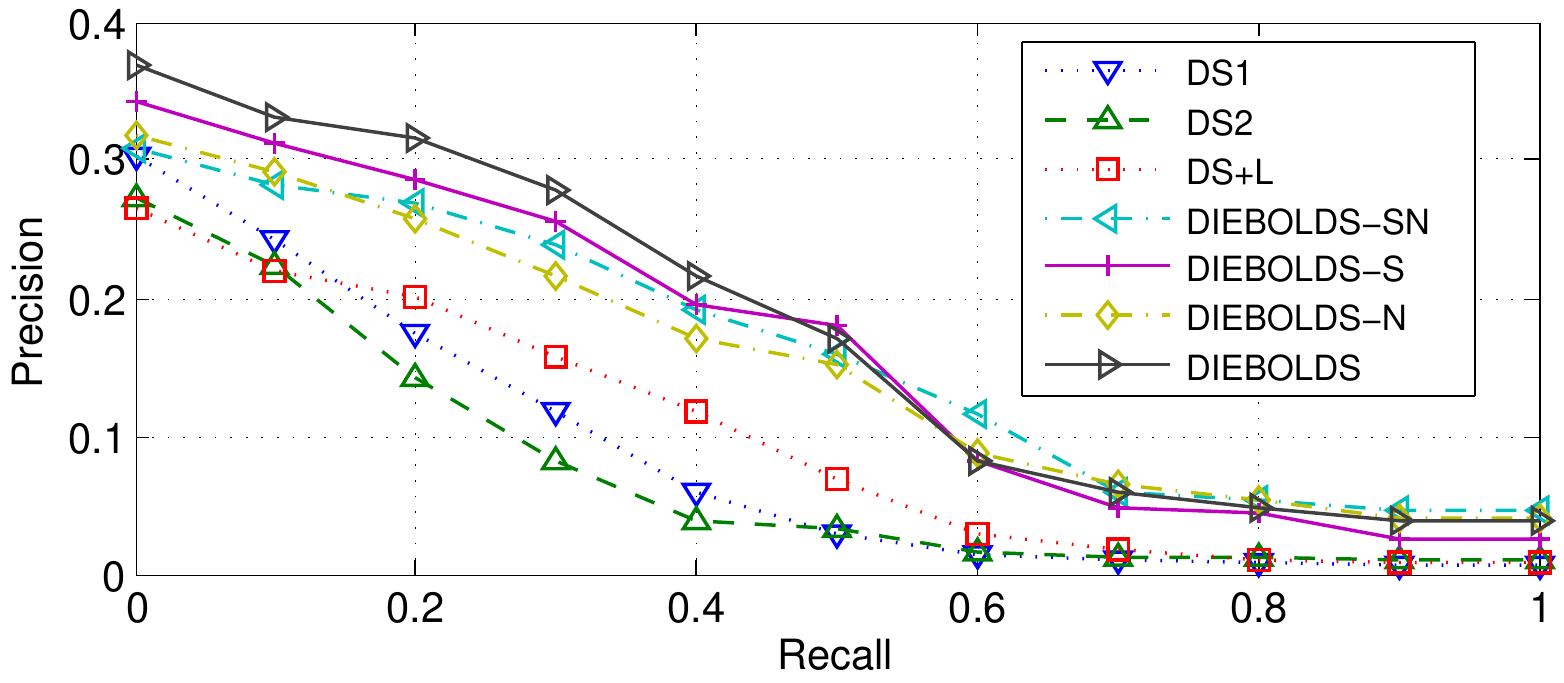}}
%\vspace{-0.1in}
\caption{Precision-recall curve of drug domain. }
\label{fig:pr-curve-drug}
\end{figure}

The precision-recall curves are given in Figures \ref{fig:pr-curve-disease}
and \ref{fig:pr-curve-drug}. We adopt the 11-point curve which is a graph plotting
the interpolated precision of an IR system at 11 standard recall levels \cite{manning2008introduction}.
In general, the top ranked results are of reasonable accuracy. For the top results,
the average precision values of DIEBOLDS are about 0.5 and 0.35.
DIEBOLDS and its variants are better than the baselines.

\subsection{Results on BioASQ Questions}

To answer some queries in BioASQ dataset, the facts from different relations need to
be combined. For example, to answer query(treatsDiseaseWithSideEffect, X , epilepsy,
spina\_bifida), the triples of used\_to\_treat and side\_effect\_of are combined in:\\
\begin{small}
\indent query(treatsDiseaseWithSideEffect, Drug, Disease, Effect)\\
\indent\indent :- used\_to\_treat(Drug, Disease), side\_effect\_of(Effect, Drug)
\end{small}
We define such rules together with the triples as input of ProPPR \footnote{https://github.com/TeamCohen/ProPPR},
to answer these queries.

We compare the triples of Freebase and DIEBOLDS pipeline.
(Output of DIEBOLDS only comes from the target corpora.)
The evaluation metrics are
Mean Average Precision (MAP) and Mean Reciprocal Rank (MRR), commonly used for question answering, as well as Recall. The results are given in Table \ref{t:bioasq_results}.
of Freebase. It shows the higher scored triples from DIEBOLDS have reasonably good accuracy. On the other hand, Freebase does not have
The recall value of DIEBOLDS is about 80\% higher than that of Freebase. It shows that DIEBOLDS returns richer knowledge.
\bing{
For both DIEBOLDS and Freebase, micro-recall value is higher than
macro-recall value. This is because they cover the answers of list questions better, which
benefits micro-recall value. The macro-recall value of DIEBOLDS is four times the value
of Freebase. The reason is that DIEBOLDS outputs more diverse facts to answer more questions.

We may extract knowledge from structured corpora as well,
so as to better meet the general goal of knowledge collection.
We did not do it here since our goal is to
investigate the use of structured corpora to help extract knowledge from target corpora.}

\begin{table}[!t]
\center{
\begin{tabular}{c|ccc}
  \hline
 & MRR & MAP & Recall \\
  \hline
  \hline
DIEBOLDS  & 0.094  & 0.092 & 0.195 \\
Freebase  & 0.025  & 0.025 & 0.109 \\
  \hline
\end{tabular}
%\end{small}

}
\caption{Results on BioASQ questions.\label{t:bioasq_results}}
\vspace{-0.2cm}
\end{table}

\section{Related Work}
%Using KBs or Wikipedia infobox as knowledge to distantly label
%training instances had been explored a few years ago \cite{wu2007autonomously,mintz2009distant}.
Distant supervision was initially employed by \citeauthor{CravenISMB99}
\shortcite{CravenISMB99} in the biomedical domain under a
different terminology, i.e. \emph{weakly labeled data}.
Then, it attracted attentions from the IE community.
%\cite{wu2007autonomously,wu2010open,mintz2009distant,riedel2010modeling,hoffmann2011knowledge,Surdeanu:2012:MML:2390948.2391003}.
\citeauthor{mintz2009distant} \shortcite{mintz2009distant}
employed Freebase to label sentences containing
a pair of entities that participate in a known Freebase relation, and
aggregated the features from different sentences for this relation.
\citeauthor{wu2007autonomously} \shortcite{wu2007autonomously,wu2010open}
also proposed an entity-centric corpus oriented method, and
employed infoboxes to label their corresponding Wikipedia articles.
DIEBOLDS employs Freebase to label entity-centric documents of particular domains.

To tolerate the noise in distantly-labeled examples, \citeauthor{riedel2010modeling}
\shortcite{riedel2010modeling} assumed that at least one of the relation mentions
in each ``bag'' of mentions sharing a pair of argument
entities which bears a relation, expresses the target relation, instead of
taking all of them as correct examples.
MultiR \cite{hoffmann2011knowledge} and Multi-Instance Multi-Label Learning (MIML)
\cite{Surdeanu:2012:MML:2390948.2391003} further improve it
to support multiple relations expressed by different
sentences in a bag. Different from them, before feeding the noisy examples into a learner,
DIEBOLDS improves the quality of training data with a bootstrapping step, which propagates
the labels in an appropriate graph. The benefit of this step is two-fold. First,
it distills the distantly-labeled examples by propagating labels through those coupling edges,
and downweights the noisy ones. Second, the propagation will
walk to other good examples that are not distantly labeled with the seeds.
In the classic bootstrapping learning \cite{riloff99Learning,agichtein00snowball,DBLP:conf/acl/BunescuM07},
small number of seed instances are used to
extract, from a large corpus, new patterns, which are used to extract more
instances. Then new instances are used to extract more patterns, in an iterative
fashion.
DIEBOLDS departs from earlier bootstrapping uses in combining label
propagation with a standard classification learner,
so that it can improve the quality of distant examples and collect new examples
simultaneously.

\section{Conclusions and Future Work}

We explored an alternative approach to distant supervision
by detection of lists in text and utilization of document structure to
overcome the weakness of distant supervision becaseu of
noisy training data. It uses distant supervision
and label propagation to find mentions
that can be confidently labeled, and uses them
to train classifiers to label more entity mentions.
The experimental results show that this approach
consistently and significantly outperforms naive
distant-supervision approaches.

For future work, one direction is to build more comprehensive
graph by integrating corpora from highly related domains.
Another worthwhile direction is to suppress the false positives,
which will significantly upgrade the overall performance.
Another approach that might be able to upgrade the performance is to use an
annotated validating page set, instead of using 10\% Freebase seeds
to automatically generate testing examples, for tuning parameters.
DIEBOLDS-SN outperforms DS+L by using the additional list information from the structured corpus.
This reminds us that using more list information of other corpora, which could be
general corpora and much larger than the target corpus, might be a worthwhile approach
to try for enhancing the extraction on the target corpus.
One might want to directly classify the drug-NP pairs on the left side
of the graph, instead of lists and mentions. This approach aggregates different
mention occurrences of the same NP, falling in the macro-reading
paradigm~\cite{Mitchell:2009:PSW:1693684.1693754}, and it might also be
a good direction to explore.

\section{Acknowledgments}
This work was funded by a grant from Baidu USA and by the NSF under research grant IIS-1250956.

%\bibliographystyle{aaai}
%\balance
%\bibliography{all}

\end{document}